\documentclass[sigconf]{acmart}
\usepackage[utf8]{inputenc}
\usepackage{multirow}
\usepackage{caption, subcaption}
\usepackage{transparent}
\DeclareUnicodeCharacter{211D}{\mathbb{R}}
\AtBeginDocument{%
  \providecommand\BibTeX{{%
    \normalfont B\kern-0.5em{\scshape i\kern-0.25em b}\kern-0.8em\TeX}}}

\begin{document}

\title{Learning Fine-grained Fact-Article Correspondence in Legal Cases}

\author{Jidong Ge}
\affiliation{%
  \institution{State Key Laboratory for Novel Software Technology, Nanjing University}
  \city{Nanjing}
  \country{China}
  \postcode{43017-6221}
}
\email{gjd@nju.edu.cn}

\author{Yunyun Huang}
\affiliation{%
  \institution{State Key Laboratory for Novel Software Technology, Nanjing University}
  \city{Nanjing}
  \country{China}}
\email{oli_yun@163.com}

\author{Xiaoyu Shen}
\authornote{Work done before joining Amazon}
\authornote{Corresponding author}
\affiliation{%
  \institution{Amazon Alexa AI}
  \city{Berlin}
  \country{Germany}
}
\email{gyouu@amazon.com}

\author{Chuanyi Li}
\authornotemark[2]
\affiliation{%
 \institution{State Key Laboratory for Novel Software Technology, Nanjing University}
  \city{Nanjing}
  \country{China}}
\email{lcy@nju.edu.cn}

\author{Wei Hu}
\affiliation{%
  \institution{State Key Laboratory for Novel Software Technology, Nanjing University}
  \city{Nanjing}
  \country{China}}
 \email{whu@nju.edu.cn}

\renewcommand{\shortauthors}{Ge and Huang, et al.}

\begin{abstract}
  Automatically recommending relevant law articles to a given legal case has attracted much attention as it can greatly release human labor from searching over the large database of laws. However, current researches only support coarse-grained recommendation where all relevant articles are predicted as a whole without explaining which specific fact each article is relevant with. Since one case can be formed of many supporting facts, traversing over them to verify the correctness of recommendation results can be time-consuming. We believe that learning fine-grained correspondence between each single fact and law articles is crucial for an accurate and trustworthy AI system. With this motivation, we perform a pioneering study and create a corpus with manually annotated fact-article correspondences. We treat the learning as a text matching task and propose a multi-level matching network to address it. To help the model better digest the content of law articles, we parse articles in form of premise-conclusion pairs with random forest. Experiments show that the parsed form yielded better performance and the resulting model surpassed other popular text matching baselines.
  Furthermore, we compare with previous researches and find that establishing the fine-grained fact-article correspondences can improve the recommendation accuracy by a large margin. Our best system reaches an F1 score of 96.3\%, making it of great potential for practical use. It can also significantly boost the downstream task of legal decision prediction, increasing the F1 score by up to 12.7\%~\footnote{Code and dataset are available at \url{https://github.com/gjdnju/MLMN}}.
\end{abstract}





\maketitle

\section{Introduction}
The legal principle for countries where the civil law system applies is that \emph{judges are obliged to respect the established statutory law articles and make their decisions based on them}. In the process of legal judgment,
properly linking each case to its related law articles is a crucial starting point, where any miss or mismatch of linkage might affect further decisions and deteriorate judicial fairness. Nonetheless, a case can contain many facts, accurately connecting all facts to their related law articles is challenging even for highly experienced legal professionals. In recent years, with the open access of big judicial data, applying AI technology to automate the search of relevant law articles has become a hot spot~\cite{LuoFXZZ17,ZhongGTX0S18,HuLT0S18,YangJZL19,chang2020dart,chang2020unsupervised}. 
However, all of them only support coarse-grained matching. All applicable law articles for a case are predicted as a whole without explaining which specific fact each article is relevant with\footnote{This is mainly attributed to the format of public legal documents. For example, China Judgments Online and European Court of Human Right usually do not explicitly provide the fine-grained correspondences, but rather explain the relationships between doctrine concepts in the later court views~\cite{smits2006elgar}}. The recommendation accuracy is also far from human performance, which greatly limits the applicability of AI systems in the legal field. We can barely trust the outputs from them and need to traverse over all facts to verify the correctness for every predicted article due to the lack of correspondence information. 

\begin{figure}
  \centering
    \begin{subfigure}{\linewidth}
        \centering
        \includegraphics[width=\linewidth]{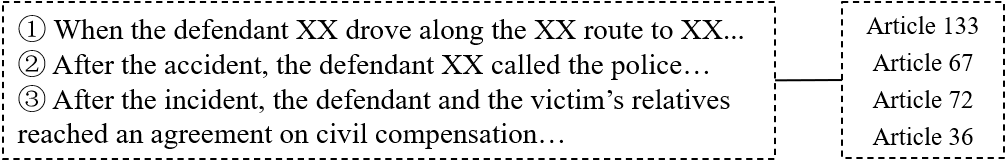}
        \caption{\small Previous coarse-grained recommendation.}
    \end{subfigure}
    \par\medskip
    \begin{subfigure}{\linewidth}
        \centering
        \includegraphics[width=\linewidth]{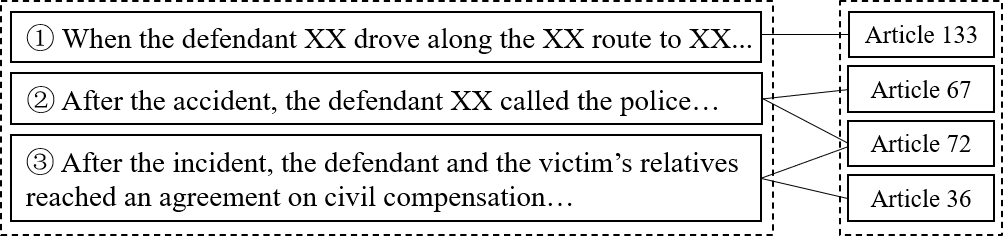}
        \caption{\small Our fine-grained recommendation.}
    \end{subfigure}
  \caption{\small Given a sequence of facts in a legal case, previous work only recommends all relevant law articles as a whole. Our work can further explain which specific fact each article is relevant with. It is able to improve (1) human interpretability, (2) recommendation accuracy and (3) downstream task of legal decision prediction.}
  \Description{Model illustration.}
  \label{fig:intro}
\end{figure}

With this in mind, this work aims to learn \emph{fine-grained fact-article} correspondences. Specifically, given any single fact from the legal case, our model will predict if a law article is applicable to it or not. The prediction can provide clear clues to help humans establish article-fact connections, or detect misused citations from a given legal document more efficiently. It can also serve as an important tool to help non-professionals interpret legal documents and thereby improve public access to justice. An example of the difference with previous research is illustrated in Figure~\ref{fig:intro}.

To enable this learning, we manually created a dataset on two domains: causing traffic accidents and intentionally injuring. For each domain, we downloaded judgment documents from China Judgments Online (CJO)\footnote{https://wenshu.court.gov.cn/} and extracted facts and law articles for annotating. We notice that in previous research, usually only charge-related law articles are extracted, while general law articles and those used for supplementary interpretation are often ignored. There are also some mistakes of law article citations even in the public legal document due to inexperienced judges, clerical or writing errors\footnote{In our annotation, we find that there are a significant amount of missing citations and over 1 mismatch per case in current datasets, detailed in Section~\ref{sec:limits}}. We aim to avoid these problems by setting the scope of the candidate articles to cover mentioned articles from all documents that we downloaded. Mistakes will be fixed beforehand.
The annotation is conducted by 15 highly qualified legal professionals who major in criminal law and takes one month to finish. The resulting dataset contains 23,306 positive fact-article pairs and 362,520 negative ones.

We regard it as a text matching task between a fact and a law article. We propose the multi-level matching network (MLMN) to tackle this problem. Our core idea is to extract the text pattern of the input text pair at multiple levels through convolution. Then the matching pattern is calculated by the attention-over-attention mechanism~\cite{cui2017attention} based on text patterns of each level. Finally, the matching patterns at all levels are identically mapped to the final prediction layer, so that the matching patterns at each level can complement each other. Besides, we parse law articles into the standard premise-conclusion format with random forest and show that it can improve the accuracy of fact-article matching. The parsed form, as being well-structured, can potentially be used to improve human interpretability and ease up the post-verification process.

Furthermore, we compare our work with previous research where only coarse-grained recommendations are provided. We find that the fine-grained fact-article annotation can significantly improve the accuracy of law article recommendation. The F1 score can reach 96.3\% (over 10\% of improvement), making it of great potential for practical use. 
It can also facilitate the downstream task of legal decision prediction (with up to 12.7\% boost in macro F1 score and 12.3\% in weighted F1 score), which further validates the advantage of including the fine-grained correspondence.

To the best of our knowledge, our work makes the first attempt at learning the \emph{fine-grained fact-article} matching to assist the legal judgment process. Our contributions can be summarized into three folds: (1) A clean manually annotated corpus with fine-grained fact-article correspondences on the traffic accident and intentional injury domains. The corpus will be made public to facilitate future research. (2) a multi-level matching network with parsed law-article information and we perform an extensive ablation study to validate the function of each component. (3) We show including the fine-grained annotation can significantly improve the accuracy of law article recommendation and the downstream task of legal decision prediction. It is also more interpretable and humans can easily verify the results by checking the correspondence.

\section{Problem Formulation}
\begin{figure}
    \centering
    \includegraphics[width=0.8\linewidth]{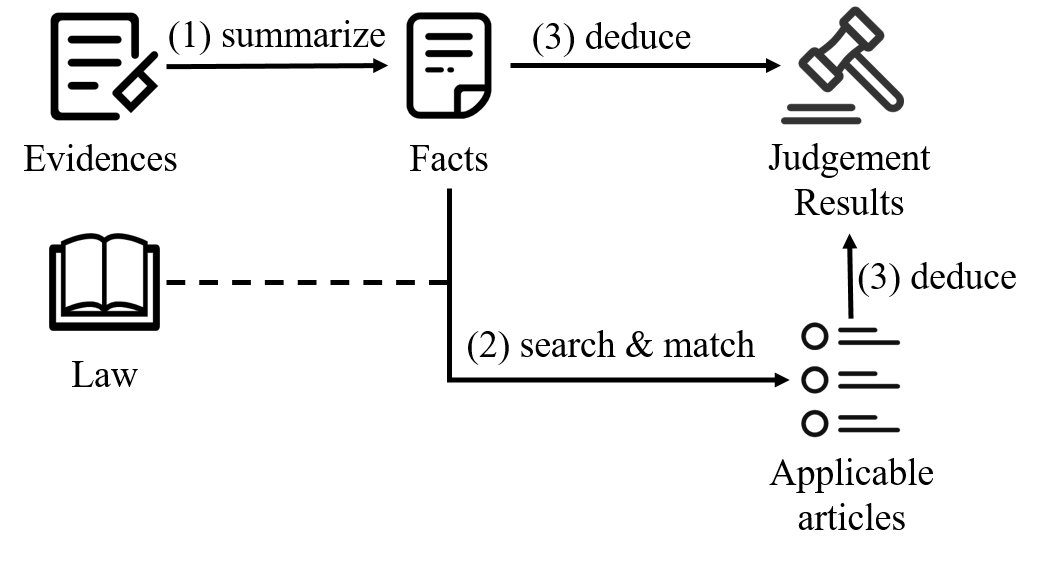}
    \caption{\small The general process of case trial in real life: (1) verify all evidences and generate the fact. (2) search for articles whose premise can match the fact. (3) combine the fact and applicable articles to deduce the judgment results.}
    \Description{Trial process}
    \label{fig:process}
\end{figure}

For trials in reality, the process of legal judgment is illustrated in Figure~\ref{fig:process}. Firstly, evidences will be presented in court and cross-examined by the parties concerned. Judges will summarize all evidences and generate the fact description. Secondly, judges will search for law articles whose premise matches with facts as the basis for the trial. Finally, according to fact descriptions and all applicable law articles, judges will make inferences and finally arrive at a series of judgment decisions such as the term of penalty, the amount of compensation and so on. In this paper, we mainly focus on the second step and regard it as a matching process between facts and law articles. Our goal is to recommend relevant law articles for a given fact description automatically. The task is crucial since it will directly affect the fairness and correctness of judgemental results.

Specifically, let $\mathcal{F}=\left(\mathbf{F^1}, \mathbf{F^2}, \cdots, \mathbf{F^t}\right)$ denote the sequence of facts in a legal case where $t$ is the number of facts in the case. We have a set of $k$ law articles $\mathcal{L}=\left(\mathbf{L^1}, \mathbf{L^2}, \cdots, \mathbf{L^k}\right)$ as candidates. Our task is to learn a functional mapping $y=M\left(\mathbf{F^i}, \mathbf{L^j}\right)$ where $M$ is a text matching network and $y \in \left\{0,1\right\}$. $y=1$ means $\mathbf{F^i}$ is relevant with law article $\mathbf{L^j}$ and vice versa.

The task is challenging considering the mismatch of contents in the law article and fact descriptions. On one hand, the definition of law articles tends to be concise and contains only a high-level abstraction of many real-life scenarios. On the other hand, fact description is relatively long, containing details of various granularity. It is also more flexible and has more diversity in the expressions. Even under the same type of cases, different habits and understanding of cases might yield totally different wording in the fact description. Besides, one fact is often related to multiple law articles. There is a many-to-many correspondence between facts and law articles which increases the difficulty of accurate matching. 

\section{Related Work}
\paragraph{\textbf{Legal Judgment Prediction}}

The task of Legal Judgment Prediction (LJP) aims to predict results of a given case automatically according to the fact description which has been studied for decades~\cite{liu2005classifying,liu2006exploring,lin2012exploiting,zhao2019unsupervised,xu2020data}. 
As an important subtask of LJP, recent researches start to pay much attention to law article recommendation to improve the interpretability and accuracy of LJP~\cite{LuoFXZZ17,ZhongGTX0S18,HuLT0S18,YangJZL19,su2018dialogue,shen2019select,shen2020neural}. \cite{LuoFXZZ17} trained a binary classifier for each law article based on SVM to predict the article-fact correlation. \cite{WangYNZZN18} proposed Dynamic Pairwise Attention Model to predict the article set for each legal case. They regarded this task as a multi-label problem and considered the relationship between articles by introducing an attention matrix learned from article definitions. \cite{YangWZSX19} proposed Recurrent Attention Network (RAN) to calculate the semantic mutual information between facts and articles repeatedly when making articles recommendation. \cite{BaoZGCX19} also jointly modeled the article recommendation task and the charge prediction task. They proposed LegalAtt which used relevant articles to filter out irrelevant information in fact based on attention mechanism. \cite{WangFNYZG19} observed the hierarchy relationship between laws and articles. They first decomposed article definitions into the residual and alignment components and then calculated the relevance scores between law and article hierarchically. \cite{ZhongGTX0S18, YangJZL19} both considered the topology among article recommendation, charge prediction and term of penalty prediction.
They modeled the dependencies among these tasks and improved performance of all tasks simultaneously. \cite{xu2020distinguish} proposed a graph neural network to automatically learn subtle differences
between confusing law articles, with which to extract compelling
discriminative features from fact descriptions. However, all of the above works focused on coarse-grained recommendation. The lack of correspondence with fact makes it hard to interpret the results and might affect the accuracy of LJP. 

\paragraph{\textbf{Text Matching}}

Text matching is a basic task in NLP. Many tasks can be formulated as text matching, such as paraphrase detection, question answering, natural language inference, etc.
The recent deep learning models for text matching can be divided into two categories: Representation-based models and Interaction-based models~\cite{guo2016deep}.
The representation-based model uses the dual-encoder Siamese architecture \cite{bromley_signature_1993}, i.e, first encodes two texts into two vectors independently, then builds a classifier on top of them to predict the output~\cite{huang_learning_2013, hu_convolutional_2014,shen2018nexus,su2020diversifying,backes2018simulating}. The independence of encoding makes it possible to store text vectors in advance for improving the online speed. However, it can only extract general text features but cannot identify useful information tailored for the matched text. 
The latter, on the contrary, considers the interaction between text pairs during the encoding process~\cite{hu_convolutional_2014, wan_deep_2016,cui2017attention,shen2017estimation,shen2019improving}. They combine the characteristics of the text pairs so that the encoder can focus on features most useful for the matching. Its main disadvantage is the sacrifice of speed and thereby hard to scale up. We adopt the interaction-based model in our case since the size candidate set is not too large in each domain.

Adding additional knowledge is also popular in NLP models~\cite{zhao2017gated,zhao2018comprehensive,qiu2020easyaug,chang2021neural}. It can be applied as a kind of feature enrichment. For example, \cite{wu_knowledge_2018} uses text categories and LDA information as knowledge and designed a knowledge gate to filter the noise in the text under the supervision of knowledge. \cite{chen_neural_2018} uses WordNet to define word relationships and used TransE~\cite{bordes_translating_2013} to get the relation vectors, which is integrated into the encoder to help the model align words more accurately. We take a similar method to incorporate the premise-conclusion information in the parsed law article.

\begin{table*}
    \caption{\small The comparison with existing corpus. ``\#Articles/Case'' is less than 1 in the first three datasets because they only extracted the final violated law article. Some cases did not violate any article specified in the dataset.}
    \centering
    \begin{tabular}{cccccccc}
        \toprule
        Dataset & Language & Data Source & \# Cases & \# Articles & \# Articles/Case & Correspondence & Manual Verify \\
        \midrule
        \citet{aletras_predicting_2016} & English & ECHR & 584 & 3 & 0.5 & no & $\times$ \\
        \citet{medvedeva_judicial_2018} & Enclish & ECHR & 11,532 & 14 & 0.5 & no & $\times$ \\
        \citet{chalkidis_neural_2019} & English & ECHR & 11,478 & 66 & 0.793 & no & $\times$ \\
        \citet{xiao_cail2018_2018} & Chinese & CJO & 2,676,075 & 183 & 1.07 & coarse & $\times$ \\
        \citet{WangYNZZN18} & Chinese & CJO & 17,160 / 4,033 & 70 / 30 & 4.1 / 2.4 & coarse  & $\times$ \\
        \midrule
        Ours & Chinese & CJO & 589 / 600 & 56 / 30 & 15.92 / 8.01 & fine & $\surd$ \\
        \bottomrule
    \end{tabular}
    \label{tab:comparison}
\end{table*}

\section{Dataset}
\subsection{Limitations of current datasets}
\label{sec:limits}
In recent years, quite a few datasets for law article recommendation have been proposed.
\citet{aletras_predicting_2016} constructed the datasets based on the cases tried by the European Court of Human Rights (ECHR). This dataset collected cases related to only 3 specific law articles. 
It was extended later by collecting more articles~\cite{medvedeva_judicial_2018, chalkidis_neural_2019}. However, they \emph{did not extract directly applicable articles for the facts at all, but rather only extracted one single violated article, if there was, referred to in the final judgment decision}. There is no direct correspondence between the fact descriptions and the final violated law article.

Compared with ECHR, CJO contains much richer resources of legal cases. CAIL2018~\cite{xiao_cail2018_2018} is a commonly used dataset for LJP. They collected judgment documents for criminal cases from CJO. However, each case is extracted with only 1 applicable law article since \emph{only articles directly related to the charge are kept}. Other articles, like the ones from the Chinese Criminal Law are ignored. 
\citet{WangYNZZN18} collected cases about Fraud and Civil action from CJO and they kept all applicable law articles cited in the original judgment documents. However, as we observed, there are significant amounts of mismatched, missing and duplicated law articles in the dataset. This can result from the judges' lack of experience or clerical errors. For example, criminal judgment documents should cite articles from laws and judicial interpretations (supplementary documents to the current law) simultaneously, but supplementary documents are ignored in many cases as they do not affect the final decision, though being important for human interpretation. There are also some wrongly written characters, abbreviations, Arabic numerals or irregular expressions in the names of the cited articles.
Many law articles cannot match any fact descriptions. The judge cites them in the document to simply support evidences or for other purposes, so they should not be included for article recommendation based on fact descriptions.
We aim to construct a high-quality dataset with fine-grained fact-article correspondences by a rigorous process of manual check to avoid the mismatched, missing or incorrect law articles. In Table~\ref{tab:comparison}, we present the comparison between our corpus and other existing ones.
\subsection{Construction Process}
We construct our dataset based on criminal judgment documents in CJO. For each case, the dataset contains a set of correspondent fact-article pairs and the final judgment decision. The construction process is as follows.

\begin{table*}
  \caption{\small The statistics of our corpus. ``\# Miss/C'' and ``\# Mismatch/C'' is the average number of missed and mismatched articles detected in the original extracted online judgment documents. We manually fix both problems in our clean corpus. Mismatches often come from articles which do not correspond to any fact, but are cited for other purposes in the document.}
  \label{tab:dataset}
  \begin{tabular}{cccccccccc}
    \toprule
    Crime & \# Cases (C) & \# Facts (F) & \# Articles (A) & \# F/C  & \# A/F & \# Words/F & \# Words/A & \# Miss/C & \# Mismatch/C \\
    \midrule
    Traffic & 589 & 4711 & 56 & 5.33 & 3.26 & 20.82 & 27.5 & 13.04 & 1.46 \\
    Injuring & 600 & 4067 & 30 & 4.73 & 1.96 & 18.47 & 14.5 & 5.72 & 1.03 \\
    \bottomrule
  \end{tabular}
\end{table*}


\textbf{Crime Domain Selection.} In Chinese Law, facts in the same crime are highly similar and law articles involved can form a limited set. Therefore, we select two crime domains with a large number of judgment documents: causing traffic accidents and intentionally injuring. For each crime, we collected nearly 600 judgment documents in CJO for follow-up research.

\textbf{Document Structure Conversion.} Since the judgment document is only semi-structured, it is difficult to obtain key elements such as facts and cited articles automatically. We adopt the method from \citet{zhuang_information_2017} to divide the original document into seven sections, which are stored in different nodes to form an XML document. Then we regard one judgment document as a unit to extract the sections of facts, law articles and term of penalty. We also unified all irregular expressions. We find that each sentence in the fact section can usually stand for an independent fact, so we use a sentence splitter to separate the section into facts. The separated facts are later passed to human annotators for post verification.

\textbf{Article Set Construction.} As mentioned above, there are quite a few missing citations in every single document. To avoid missing to the largest extent, when annotating the correspondence, we use all articles that have ever been mentioned in \emph{all extracted documents} (in the same crime domain) as the candidate set. Namely, if one article is missing in one document, but mentioned in another document, we can still bring it back and assign it to the fact it corresponds to.

\textbf{Judgment Decision Extraction.} We also extract the judgment decision in each document so that we can see if the correspondence can help predict the decision. We regard this prediction as multi-class classification. There are too many classes it can be involved with, so we divided all judgment results into 5 classes: exempt from criminal punishment, criminal detention, fixed-term imprisonment of not more than 1 year, 1 - 3 years and not less than 3 years. 

\textbf{Fact-Article Correspondence Annotation.} Since the language used in legal documents is highly domain specific, we recruited fifteen highly qualified annotators who major in criminal law. As a preliminary guideline, we define a law article is relevant with a fact if one of the following conditions happen: (1) If a fact summarizes some basic circumstances which can match one of the premises in a law articles, since the law can provide clues on deciding the consequence. (2) If a fact includes some post-incident behaviors such as compensation, compromise which matches one of the conclusions in law articles, since the law can provide basis to judge these behaviors. (3) If a law article provides definitions such as surrendering, hit-and-run, etc for circumstances in the fact, since 
they can help define the behaviors in the fact. All annotators were divided into five groups and each group was assigned nearly 240 judgment documents. In each group, two of them were responsible for the first stage of annotating. Basically, for each fact, the annotator needs to mark all relevant articles from the article set. If there is a disagreement between them, a third annotation will be collected and the final results are obtained via majority voting. We find that an agreement score of 97.5\% for the first-stage annotation of fact-article pairs, the remaining 2.5\% of the pairs required one more voting process to determine the labeling results, usually due to the annotators' carelessness or the cases' complexity. In order to improve the effectiveness of annotation, we have designed an annotation system for this task. The system is designed to ensure: (1) Co-existence of relevant law articles. For the pair of articles which have strong correlation, such as the article of surrendering and the article for supplementary interpretation of surrendering, they must be selected together. (2) mutex of irrelevant law articles. The law articles involved in the same charge can be mutually exclusive. For example, the articles describing different actions which lead to different types of punishment cannot be selected at the same time. In the end, the whole annotating process takes one month to finish.

\textbf{Corpus Statistic.} The statistics of our dataset is in Table~\ref{tab:dataset}. Comparing our annotation results with the original judgment documents, we found that the missing citations are mostly judicial interpretations. The mismatches are usually because we only extracted three sections of contents in the original document and some law articles related to the filtered contents cannot match with any facts.

\section{Model}
Text matching models are usually representation-based or interaction-based. In this section, we will first introduce some classic models in the field of text matching, then present our proposed MLMN.

\subsection{Representation-based models}
\paragraph{\textbf{ARC-I}} ARC-I~\cite{hu_convolutional_2014} generates distributed representations of two texts based on CNN respectively, and then concatenates the two together to predict the matching result by Multi-Layer Perceptron. The disadvantage of ARC-I is that it does not make interaction between texts until they finish encoding, so the text representation cannot incorporate the information of the other text. 

\paragraph{\textbf{MV-LSTM}} MV-LSTM~\cite{wan_deep_2016} uses bidirectional LSTM (Bi-LSTM) to generate sentence-level representations at each position. Then it calculates the interaction matrix between the two texts' representations and selects top k matching features for the final prediction.

\subsection{Interaction-based models}
\paragraph{\textbf{ARC-II}} In view of the drawback of ARC-I, \citet{hu_convolutional_2014} proposed ARC-II which continues convolution based on the interaction of two texts after encoding. However, it is worth noting that ARC-II is less salient than ARC-I when sentences have deep grammatical structures. How to balance the relationship between representation and interaction is non-trivial.  

\paragraph{\textbf{Gaussian-Transformer}} Transformer~\cite{vaswani2017attention} is an emerging text encoder which relies on the attention mechanism completely to model dependencies in texts and realizes the parallelization of computation. Gaussian-Transformer~\cite{guo2019gaussian} adds several interaction blocks after encoding blocks of the original Transformer. It merges interaction information into both texts by inter-attention. 

\paragraph{\textbf{Match-Pyramid}} Different from the above models which capture interaction information based on sentence-level representation, Match-Pyramid~\cite{pang2016text} starts from the word level to obtain higher level interaction information layer by layer through convolution and uses interaction information in the highest level to predict matching results. They pointed out that the interaction information of texts is hierarchical and high level interaction information can be generated by the combination of lower level interactions. However, we noticed that Match-Pyramid only make prediction based on the final interaction information. In this way, the useful information at the lower layers may be lost after multiple convolutions which can lead to a decline when predicting the final results. Inspired by this model, we use neural networks to extract text patterns and \emph{make interactions at different levels}. At last, we use interaction information at all layers to predict the matching results.

\paragraph{\textbf{Attention-over-Attention}} As for the calculation of interaction information, there are some commonly used methods, including cosine, indicator function and dot product. \citet{pang2016text} show that dot product can well highlight important words when doing text matching, so we also use this method to calculate the correlation matrix between texts. For the alignment mechanism, most matching models~\cite{yang2019simple, chen_neural_2018} adopt unidirectional attention to calculate the alignment results in two directions sequentially and then make further reasoning respectively. We refer to the attention-over-attention mechanism~\cite{cui2017attention} which compresses the result in one of the alignment directions firstly, and then merges it into the other alignment result, changing the bidirectional alignment into ``unidirectional''. This mechanism not only keeps the information in both of two directions, but also reduces subsequent computing greatly.

\begin{figure}[ht]
    \centering
    \includegraphics[width=\linewidth]{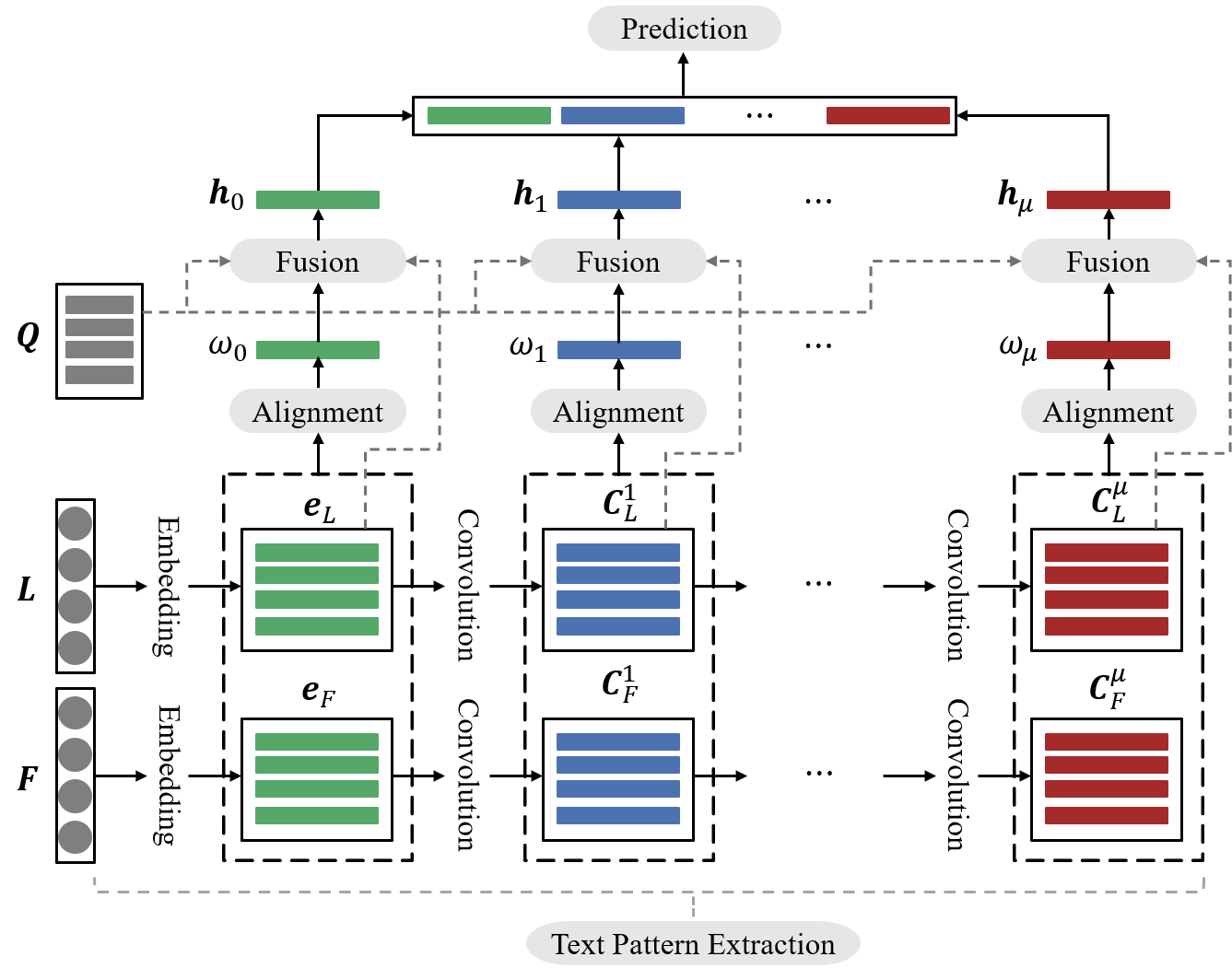}
    \caption{\small The overall architecture of MLMN.}
    \label{fig:architecture}
\end{figure}

\subsection{Multi-Level Matching network (MLMN)}
We propose multi-level matching network (MLMN), which combines the hierarchical encoding idea from match-pyramid and the interaction mechanism from attention-over-attention (AoA). It contains a multi-layer encoder and every layer interacts with the paired text through attention-over-attention. The architecture is illustrated in Figure~\ref{fig:architecture} and consists of four parts: text pattern extraction, alignment, fusion and prediction.
Here we represent the fact to be matched as $\mathbf{F}=\left(f_1, f_2, \cdots, f_m\right)$, the law article as $\mathbf{L}=\left(l_1, l_2, \cdots, l_n\right)$ which are both word sequences. $m$ and $n$ are the number of the facts and the law articles respectively. 

\paragraph{\textbf{Text pattern extraction}} We found that the CNN encoder leads to superior performance compared with LSTM and Transformer (See Sec~\ref{sec:ablation}), so we use one-dimensional convolution operation to extract text patterns. Before performing the convolution operation, the padding operation is performed on both ends of the input text to ensure the length of the output is consistent. Suppose that $\mathbf{x}_{i} \in \mathbb{R}^{d}$ represents the $d$-dimensional word vector corresponding to the $i$-th word in the sequence. We represent the sequence of length $n$ as:
\begin{equation}
\mathbf{x}_{1: n}=\mathbf{x}_{1} \oplus \mathbf{x}_{2} \oplus \ldots \oplus \mathbf{x}_{n}
\end{equation}
where $\oplus$ is the concatenation operator. Then we use the one-dimensional convolution operation to produce a new feature with a filter $\mathbf{w} \in \mathbb{R}^{h \times d}$ which is applied to a window of $h$ words:
\begin{equation}
c_{i}=\mathbf{w} \cdot \mathbf{x}_{i: i+h-1}+b \quad i=1,2,\dots,n-h+1
\end{equation}
where $\mathbf{x}_{i: i+h-1}$ denotes the concatenation of words $\mathbf{x}_{i},\mathbf{x}_{i+1},\dots,$ $\mathbf{x}_{i+h-1}$ and $c_{i}$ is the generated feature. This filter is applied to each possible window of words, i.e., $\mathbf{x}_{1: h}, \mathbf{x}_{2: h+1}, \dots, \mathbf{x}_{n-h+1: n}$. Afterwards, we obtain a feature map $\mathbf{c} \in \mathbb{R}^{n-h+1}$ with this filter:
\begin{equation}
\mathbf{c}=\left[c_{1}, c_{2}, \ldots, c_{n-h+1}\right]
\end{equation}
In this way, we obtain the text pattern $\mathbf{C}^{i}_{F} \in \mathbb{R}^{m \times {\lambda}_{i}}$ and $\mathbf{C}^{i}_{L} \in \mathbb{R}^{n \times {\lambda}_{i}}$ for fact and law article respectively at the $i$-th layer where ${\lambda}_{i}$ represents the number of filters at this layer. When $i=0$, the text patterns are their word embedding obtained from our pre-trained embedding model. With the increase of $i$, the extracted feature is expected to be more abstract.


\paragraph{\textbf{Alignment}} The inputs of this stage are text patterns extracted from each layer. For a pair of fact and article's text patterns, the alignment layer calculates the correlation between them to obtain the alignment results. Here we adopt AoA as the alignment mechanism.
For text patterns $\mathbf{C}_{F}$ and $\mathbf{C}_{L}$ (the indicator of layer number is omitted here), we calculate the correlation matrix $\mathbf{M} \in \mathbb{R}^{m \times n}$ between fact and law article by matrix multiplication. Let $\mathbf{M}_{i,j}$ denote the correlation score between the text representation at the $i$-th position of fact and that at the $j$-th position of law article. Then we calculate the weights of each position from both two alignment directions with attention mechanism. Concretely,
\begin{equation}
\mathbf{M}=\mathbf{C}_{F} \mathbf{C}_{L}^\mathrm{T}
\end{equation}
\begin{equation}
    \mathbf{\alpha}_{i, j}=\frac{\exp \left(\mathbf{M}_{i, j}\right)}{\sum_{k=1}^{n} \exp \left(\mathbf{M}_{i, k}\right)}
\end{equation}
\begin{equation}
    \mathbf{\beta}_{i, j}=\frac{\exp \left(\mathbf{M}_{i, j}\right)}{\sum_{k=1}^{m} \exp \left(\mathbf{M}_{k, j}\right)}
\end{equation}
where $\alpha_{i, j}$ represents the weight of $j$-th position in law article when it aligns all positions of fact. The larger the weight is, the more relevant and the more important this position is. The calculation of $\beta_{i, j}$ is similar to $\alpha_{i, j}$, but the alignment direction is opposite. In this way, we get two weight matrices $\alpha \in \mathbb{R}^{m \times n}$, $\beta \in \mathbb{R}^{m \times n}$. Then we need to compress one of the results of two directions. Considering the diversity of facts and the fixity of law articles, what we compress is the fact weight matrix $\beta$, that is the alignment result that we use fact to align article. The compression strategy is to calculate the cumulative weight of each position in the fact and then we obtain the overall weight vector of fact $\nu \in \mathbb{R}^{m}$ where 
\begin{equation}
    \nu_{i}=\sum_{k=1}^n \beta_{i,k} \quad i=1,2,\dots,m
\end{equation}
represents the overall weight of the $i$-th position in the fact. At last, the compression result is merged into the other alignment result. We conduct matrix multiplication between each column of the article weight matrix and the overall weight vector of fact:
\begin{equation}
    \omega_i = \nu \alpha_i \quad i=1,2,\dots,n
\end{equation}
which is the final comprehensive weight vector $\omega \in \mathbb{R}^n$ of the law article's text pattern at this layer.

\begin{figure}[t]
    \centering
    \includegraphics[width=\linewidth]{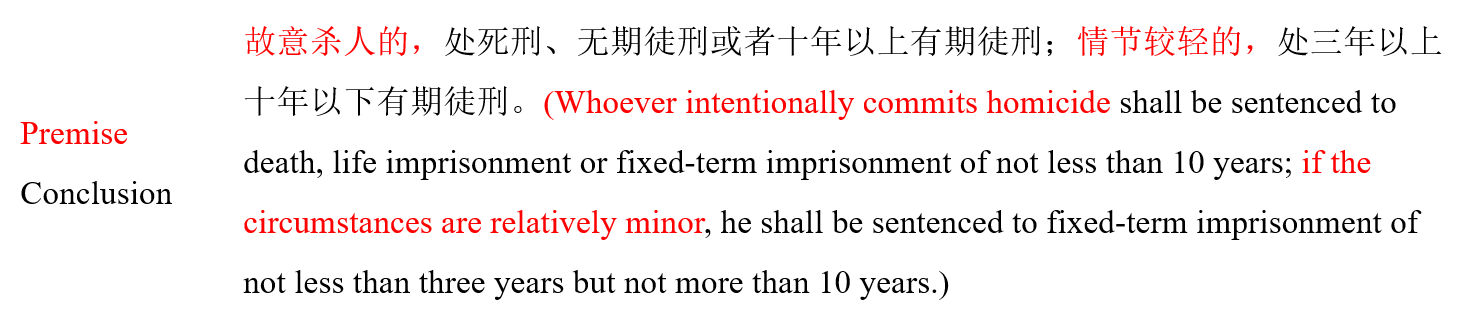}
    \caption{\small law articles parsed into premise -conclusion pairs.}
    \label{fig:knowledge}
\end{figure}

\paragraph{\textbf{Parsing law article}} We find that most of these articles follow a form of premise-conclusion. Take Article 232 of the Criminal Law as an example, this article contains two pairs of premises and conclusions as shown in Figure~\ref{fig:knowledge}. Based on this observation, we use regular expressions to split each law article into multiple clauses and summarize text rules of premises and conclusions. Finally, we use random forest as our binary classifier and train it on our annotated corpus. Through this classifier, given a law article $\mathbf{L}=\left(l_1, l_2, \cdots, l_n\right)$, we can predict the corresponding class $\mathbf{Q}=\left(\mathbf{q}_{1},\mathbf{q}_{2},\dots,\mathbf{q}_{n}\right)$, where $\mathbf{q}_{i} \in \mathbb{R}^{2}$ indicates whether the clause of $i$-th word belongs to premise or conclusion. If this word is in the premise clause, then $\mathbf{q}_{i}=[1,0]$, otherwise $\mathbf{q}_{i}=[0,1]$.

\paragraph{\textbf{Fusion and prediction}} After the above steps, we have text patterns $\mathbf{C}_{F}^{i}$, $\mathbf{C}_{L}^{i}$, the article weight with alignment information $\omega_{i}$ in different layers and the external knowledge of law articles $\mathbf{Q}$. The fusion process of above information to obtain the matching pattern in $i$-th layer is as follows.
Firstly, we calculate the new expression of law article based on the comprehensive weights:
\begin{equation}
    \mathbf{r}_{i}=\omega_{i} \ast \mathbf{C}_{L}^{i}
\end{equation}
where $\ast$ represents element-wise multiplication here. Then we concatenate it with the additional knowledge $\mathbf{Q}$ and send it to a fully-connected network $G_{1}^{i}$. Suppose that the output dimension of this layer is $t$, we get $\mathbf{u}_i \in \mathbb{R}^{n \times t}$. Then we apply the max pooling operation on its last dimension to keep the most important features as the matching pattern:
\begin{equation}
    \mathbf{u}_{i} = G_{1}^{i}\left(\left[ \mathbf{r}_i, \mathbf{Q} \right] \right)
\end{equation}
\begin{equation}
    \mathbf{h}_{i} = max\_pool\left(\mathbf{u}_{i}\right) \quad i=0,1,\dots
\end{equation}
In this way, we get the matching pattern $\mathbf{h}_{i} \in \mathbb{R}^{n}$ at each layer. To make the final prediction, we concatenate all matching patterns together. We use the fully-connected network and Softmax to calculate the probability distribution of the match results. That is
\begin{equation}
\mathbf{p}=G_{2}\left(\left[\mathbf{h}_{0}, \mathbf{h}_{1}, \dots \right] \right)    
\end{equation}
where $G_{2}$ is a fully connected neural network with a single hidden layer, and the activation function of this hidden layer is ReLU. When the probability of match is greater than a given threshold, we will recommend this law article to the corresponding fact. 

\section{Experiments and Analyses}

\subsection{Experimental Settings}
\paragraph{\textbf{Data Preprocess}} We use Jieba\footnote{https://github.com/fxsjy/jieba} for Chinese word segmentation and POS tagging. 
Since the parsed law-article information is of word granularity, we need to project this information of each clause into each word in them. Therefore, unlike performing word segmentation in facts directly, we first split the law articles into the list of clauses and use a pre-trained random forest classifier to predict if each clause is a premise or conclusion. Then we do word segmentation and get the parsed law-article information with the same length for each clause. Since law articles are relatively concise and have fewer noisy words, we only delete stop words from the segmentation results. Finally, the segmentation and the parsed information obtained from each clause are concatenated.

\paragraph{\textbf{Implementation Details}} We split all the judgment documents into 80\% for training, 10\% for validation and 10\% for testing for each crime domain. Based on all the documents we collected, we use the Gensim toolkit to train the Word2Vec model with CBOW~\cite{mikolov2013efficient}. The dimension of the word vector is set to 128. The input length of facts and articles is fixed as 50 words. The number of convolution kernels for all layers is 128. The size of convolution kernel increases with the number of layers $\left(2, 4, \dots \right)$. The threshold for determining the matched category is set to 0.6. When training the model, dropout~\cite{hinton2012improving} and early-stopping~\cite{caruana2001overfitting,su2020moviechats} are used to alleviate the over-fitting phenomenon. We use cross-entropy loss function and the Adam optimizer~\cite{kingma2014adam} to optimize parameters. Considering the unbalance between positive and negative samples, we randomly selected part of negative samples with a ratio of 1:12 and 1:5 respectively for the crime of causing traffic accidents and intentionally injuring. The baseline models are mainly implemented based on the MatchZoo toolkit~\cite{guo2019matchzoo}. As for our proposed model, we set the number of layers to 3 for both crime domains.

\subsection{Main Results}

\begin{table}
    \caption{\small Performance comparison with baseline models. Starred values are significantly better than all baselines with $p=0.05$.}
    \centering
    \resizebox{\linewidth}{!}{
    \begin{tabular}{ccccccc}
        \toprule
        \multirow{2}{*}{Model} & \multicolumn{3}{c}{Traffic} &  \multicolumn{3}{c}{Injuring} \\
         & P & R & F1 & P & R & F1 \\
        \midrule
        ARC-I & $0.862^{\transparent{0} *}$ & $0.853^{\transparent{0} *}$ & $0.857^{\transparent{0} *}$ & 0.929 & $0.853^{\transparent{0} *}$ & $0.890^{\transparent{0} *}$ \\
        MV-LSTM & $0.877^{\transparent{0} *}$ & $0.807^{\transparent{0} *}$ & $0.841^{\transparent{0} *}$ & 0.942 & $0.830^{\transparent{0} *}$ & $0.882^{\transparent{0} *}$ \\
        G-Transformer & $0.611^{\transparent{0} *}$ & $0.593^{\transparent{0} *}$ & $0.601^{\transparent{0} *}$ & 0.848 & $0.776^{\transparent{0} *}$ & $0.810^{\transparent{0} *}$ \\
        Match-Pyramid & $0.882^{\transparent{0} *}$ & $0.808^{\transparent{0} *}$ & $0.843^{\transparent{0} *}$ & 0.934 & $0.814^{\transparent{0} *}$ & $0.870^{\transparent{0} *}$ \\
        ARC-II & $0.858^{\transparent{0} *}$ & $0.892^{\transparent{0} *}$ & $0.875^{\transparent{0} *}$ & 0.931 & $0.870^{\transparent{0} *}$ & $0.900^{\transparent{0} *}$ \\
        AoA & $0.887^{\transparent{0} *}$ & $0.869^{\transparent{0} *}$ & $0.878^{\transparent{0} *}$ & 0.929 & $0.877^{\transparent{0} *}$ & $0.902^{\transparent{0} *}$ \\
        \midrule
        MLMN & $0.896^*$ & $0.901^{\transparent{0} *}$ & $0.899^*$ & $\mathbf{0.944}$ & $0.885^{\transparent{0} *}$ & $0.914^*$ \\
        + parsed & $\mathbf{0.905}^*$ & $\mathbf{0.913}^*$ & $\mathbf{0.909}^*$ & 0.940 & $\mathbf{0.907}^*$ & $\mathbf{0.923}^*$ \\
        \bottomrule
    \end{tabular}}
    \label{tab:main_results}
\end{table}

\begin{table}
    \caption{\small Performance comparison of using different encoders.}
    \centering
    \resizebox{\linewidth}{!}{
    \begin{tabular}{ccccccc}
        \toprule
         \multirow{2}{*}{Encoder} & \multicolumn{3}{c}{Traffic} &  \multicolumn{3}{c}{Injuring} \\
         & P & R & F1 & P & R & F1 \\
        \midrule
        CNN & 0.893 & $\mathbf{0.893}^*$ & $\mathbf{0.893}$ & $\mathbf{0.918}$ & $\mathbf{0.902}^*$ & $\mathbf{0.910}$ \\
        LSTM & $\mathbf{0.894}$ & $0.874^{\transparent{0} *}$ & 0.884 & 0.915 & $0.890^{\transparent{0} *}$ & 0.903 \\
        Transformer & 0.618 & $0.440^{\transparent{0} *}$ & 0.514 & 0.854 & $0.750^{\transparent{0} *}$ & 0.799 \\
        \bottomrule
    \end{tabular}}
    \label{tab:encoder}
\end{table}

The comparison with baseline models on two crime domains is shown in Table~\ref{tab:main_results}. It can be observed that MLMN achieves better performance than all of baselines. By adding the parsed information of law articles, the performance of our model is further improved. The best systems achieve an F1 score of over 90\% on both domains, making it of great potential for practical use.

Among all baselines, the interaction-based ARC-II and AoA outperforms, as expected representation-based ARC-I and MV-LSTM for being able to attend to each other before obtaining the vector representation. It verifies the advantage of mining the dependency of local features between facts and law articles. AoA performs slightly better than ARC-II, but the difference is marginal. Match-Pyramid does not perform well here. We suppose that this model starts interaction from the word level which relies on the quality of word embedding to a great extent. With the increase of convolutional layers, effective information at lower levels may be lost, which limits the performance of this model. Since Gaussian-Transformer does not have publicly available codes, we have tried our best to reproduce this paper. The model we implementing does not achieve good results. One possible reason is that our corpus does not have enough samples which limits the model's performance. Transformer-based encoders, as known, are highly dependent on large quantities of training data to show the benefits.

\subsection{Ablation Study}
\label{sec:ablation}
To have an in-depth understanding of MLMN, we perform a set of ablation studies to test our model under different settings.

\begin{figure}
    \begin{subfigure}{0.7\linewidth}
        \centering
        \includegraphics[width=\linewidth]{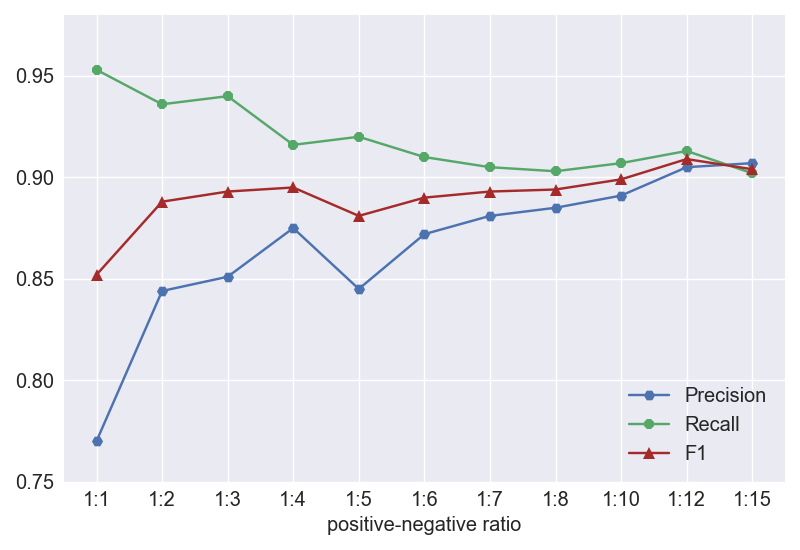}
        \caption{\small The crime of causing traffic accidents.}
    \end{subfigure}
    \begin{subfigure}{0.7\linewidth}
        \centering
        \includegraphics[width=\linewidth]{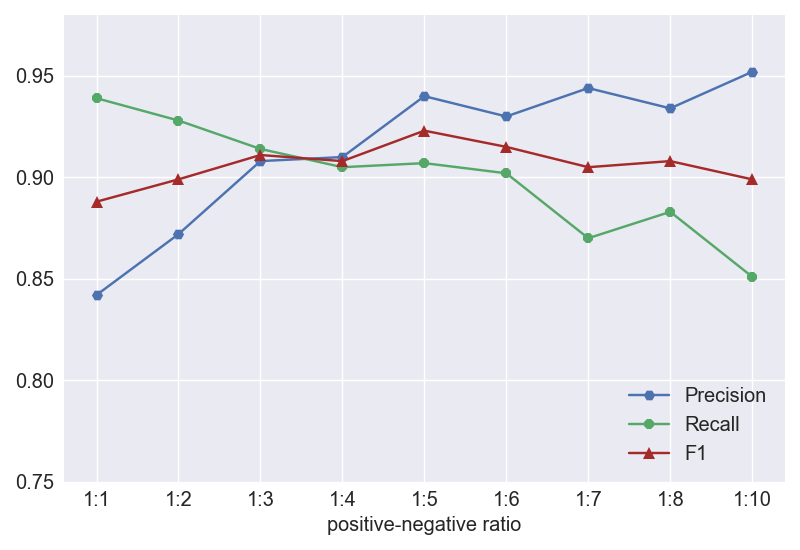}
        \caption{\small The crime of intentionally injuring.}
    \end{subfigure}
    \centering
    \caption{\small MLMN Performance under different positive-negative ratio. F1 is rather stable but there is a clear precision/recall trade-off.}
    \label{fig:negtive_ratio}
\end{figure}

\paragraph{\textbf{Positive-Negative Ratio}}
Since our dataset is imbalanced, we adopt the strategy of randomly selecting part of negative samples in each batch data while training. In Figure~\ref{fig:negtive_ratio}, we illustrate the performance of MLMN under different positive-negative ratios. It can be observed that in both two crimes, as the number of negative samples continues to increase, precision continues to rise, while recall continues to fall. This is understandable as the model has to pay more attention on distinguishing negative samples. We can tune the ratio to suit our need in practice. In this work, we have selected 1:12 and 1:5 which lead to the highest F1 score for two crime domains respectively.

\paragraph{\textbf{Text Encoder}} In this experiment, we test the effects of using different types of text encoders to generate text patterns. We consider three popular encoding mechanisms: CNN, Bi-LSTM and Transformer. For CNN and Bi-LSTM, we both adopt a two-layer architecture. For Transformer, we use two encoder blocks. The experimental results are shown in Table~\ref{tab:encoder}. The performance of Bi-LSTM is slightly worse than CNN. The results of the Transformer is very bad. This might due to the same reason as we mentioned above. The fully-attention Transformer has difficulty being well trained under so few training samples.

\begin{figure}[t]
    \centering
    \begin{subfigure}{0.45\linewidth}
        \centering
        \includegraphics[width=\linewidth]{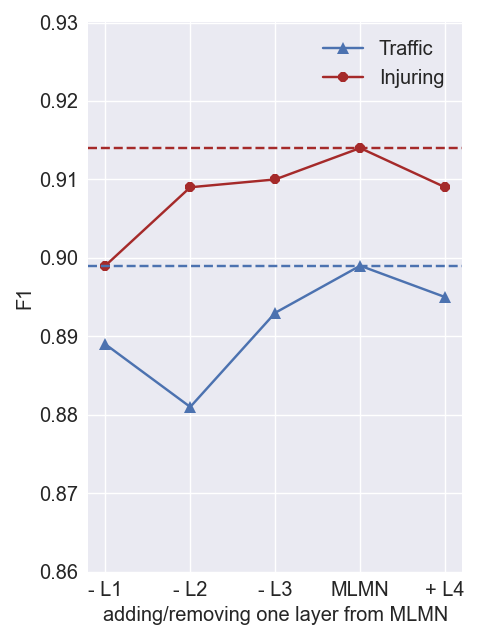}
        \caption{\small}
        \label{fig:layers_a}
    \end{subfigure}
    \hspace{0.2cm}
    \begin{subfigure}{0.45\linewidth}
        \centering
        \includegraphics[width=\linewidth]{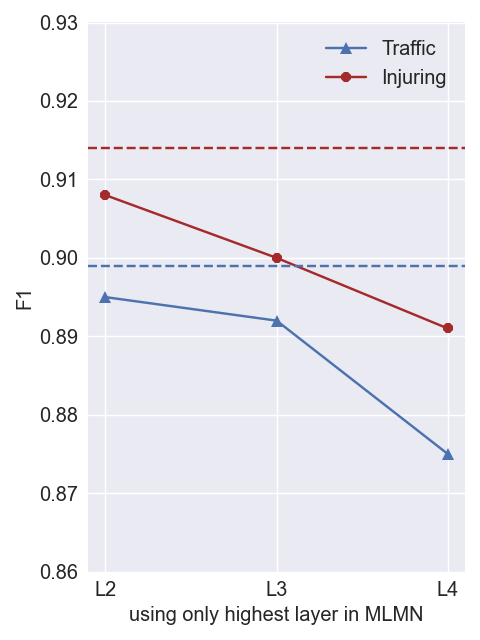}
        \caption{\small}
        \label{fig:layers_b}
    \end{subfigure}
    \caption{\small (a) The F1 score of MLMN after removing (-L$N$) or adding (+L$N$) certain layer. Removing or adding any layer in our model hurts the performance. (b) The F1 score of only using the highest-layer matching pattern to make prediction (with varying number of layers). Results for only using L$1$ is not displayed for visualization issue. Its score is very low with only static word embeddings.}
    \label{fig:layers} 
\end{figure}

\paragraph{\textbf{Multi-Level}} With the number of convolutional layers deepening, the encoded information of corresponding matching patterns becomes higher-level. In order to verify the necessity of using multi-level information, we conduct experiments with different layers. The experimental results are shown in Figure~\ref{fig:layers}. We can observe that if we delete any one layer in MLMN, the performance decreases in different degrees. When using only the last layer for prediction, the performance decays when more layers are added. The high-level text pattern can summarize the whole text better, but useful information in low levels may be lost after layers of convolution which may lead to lower performance as shown in Figure~\ref{fig:layers_b}. Furthermore, with the increment of layers, the amount of parameters in model is also increased, which can make training process harder. Nonetheless, we can see MLMN clearly outperforms using only the last layer for prediction (all models in Figure~\ref{fig:layers_b}). Our design of multi-level interaction is beneficial for detecting the correspondence.

\begin{figure}[t]
    \centering
    \begin{subfigure}{0.45\linewidth}
        \centering
        \includegraphics[width=\linewidth]{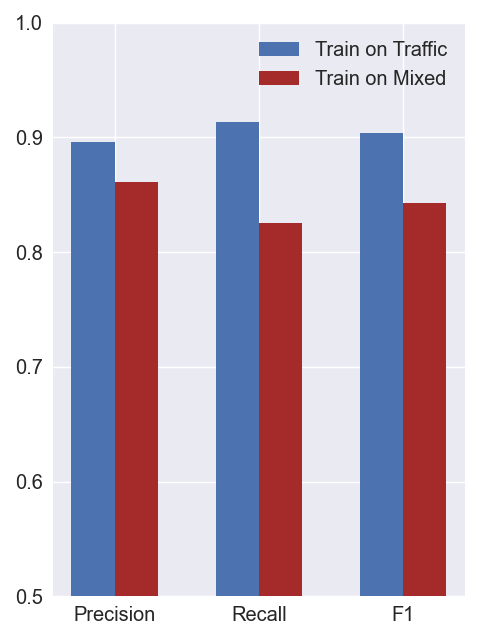}
        \caption{\small Test on Traffic.}
    \end{subfigure}
    \hspace{0.2cm}
    \begin{subfigure}{0.45\linewidth}
        \centering
        \includegraphics[width=\linewidth]{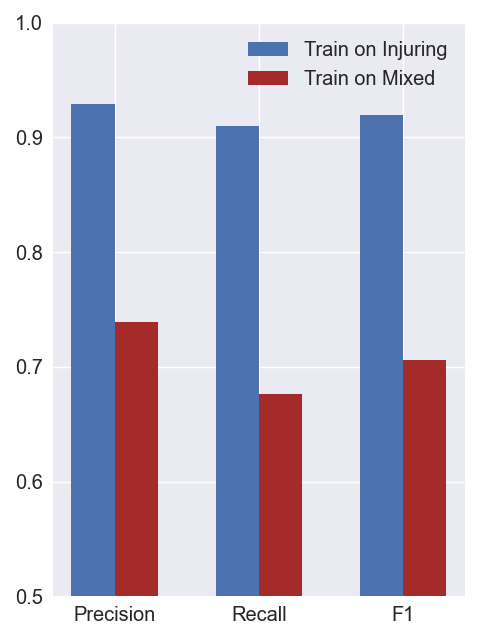}
        \caption{\small Test on Injuring.}
    \end{subfigure}
    \caption{\small Performance with mixed/separated Training.}
    \label{fig:crime}
\end{figure}

\paragraph{\textbf{Mixed or Separate}} In order to see if it is better to train on each crime domain separately or using a single system trained on the mixed dataset, we compare their performance and shown in Figure~\ref{fig:crime}. After training on mixed data, the performance on the test set of traffic accidents is much better than intentionally injuring due to the imbalance in the total number of samples of these two crimes. The performance of training models separately on each crime domain is better than training on the mixed dataset. We suppose that different crime domains have different features of cases and those frequently cited articles are also different. When the data of different crimes are mixed for training, the model not only needs to learn model semantic connections between facts and law articles, but also needs to learn the fact expression patterns of different crimes, which increases the difficulty of training, making the model broad yet shallow. Training separately encourages the model to learn the mapping specifically for the crime domain.

\begin{figure}
    \centering
    \begin{subfigure}{\linewidth}
        \centering
        \includegraphics[width=\linewidth]{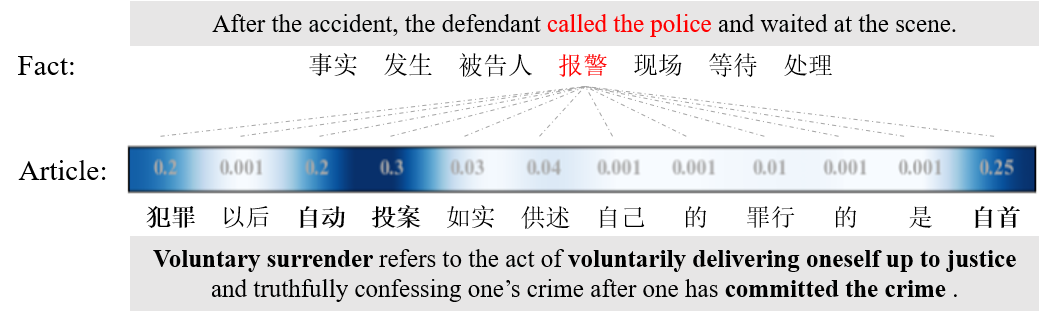}
        \caption{\small Use fact to align with law article.}
        \label{fig:att_a}
    \end{subfigure}
    \par\medskip
    \begin{subfigure}{\linewidth}
        \centering
        \includegraphics[width=\linewidth]{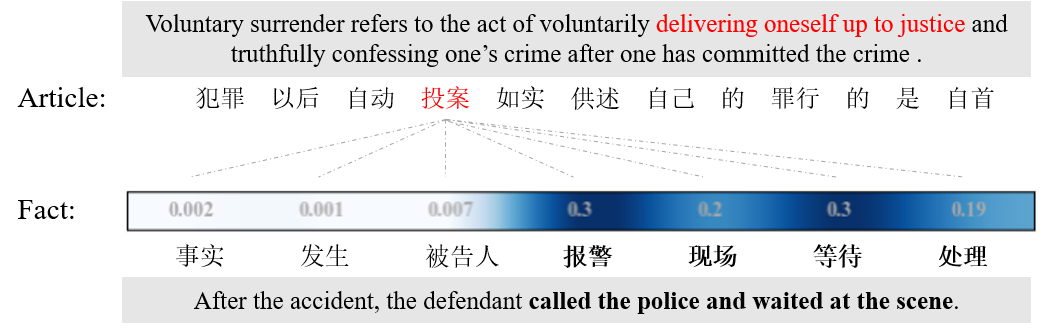}
        \caption{\small Use law article to align with fact.}
    \end{subfigure}
    \caption{\small An illustration of aligning texts on different directions. The text in shallow is the translation of the original text. The darker color means higher attention weights with the word marked in red.}
    \label{fig:attention}
\end{figure}

\paragraph{\textbf{Direction of Attention-over-Attention}} Figure~\ref{fig:attention} gives an example of aligning texts based on different directions. Taking the fixed content of law articles into account, we compress the fact weight matrix in AoA. We can see the model indeed is able to align corresponding words (e.g., ``call the police'' is aligned with ``voluntary surrender'' in Figure~\ref{fig:att_a}. In order to study the effect of the fusion direction, we designed two comparative experiments: (1) using comprehensive weights of fact and that of article, which represent integrating article into fact and integrating fact into article respectively. (2) replacing the average operation in the original mechanism with the summation operation in our framework. The reason for this change is to be consistent with the subsequent weighted sum. Experiments are also designed to verify the effectiveness of this modification. All models here did not add any external knowledge. Results are shown in Table~\ref{tab:AoA} which are consistent with expectations. We suppose that facts are always related to one or several specific article’s conditions. When we use comprehensive weight of the article for prediction, the model is easier to learn the weight distribution patterns of each article in current crime. On the contrary, due to the diversity and complexity of facts, even if two comprehensive weights of fact are calculated by two similar facts and one same article, there is still a larger difference between them which makes the performance not as good as the former. 

\begin{table}
    \caption{\small Comparison of different fusion direction and compress operation of AoA. The arrow denotes the direction of fusion. ``Compress Operation'' represents the operation of compress matrix before integrating. F1 score is reported.}
    \centering
    \begin{tabular}{cccc}
        \toprule
        Direction & Compress Operation & Traffic & Injuring \\
        \midrule
        Article $\rightarrow$ Fact & Sum & $0.886^{\transparent{0} *}$ & $0.874^{\transparent{0} *}$ \\
        Fact $\rightarrow$ Article & Sum & $\mathbf{0.899}^*$ & $\mathbf{0.914}^*$ \\
        Fact $\rightarrow$ Article & Avg & $0.884^{\transparent{0} *}$ & $0.905^{\transparent{0} *}$ \\
        \bottomrule
    \end{tabular}
    \label{tab:AoA}
\end{table}

\subsection{Comparison between Coarse/Fine-grained}

\begin{table}[]
    \centering
    \caption{\small Comparison of coarse/fine-grained recommendation.}
    \resizebox{\linewidth}{!}{
    \begin{tabular}{ccccccc}
        \toprule
        \multirow{2}{*}{Correspondence} & \multicolumn{3}{c}{Traffic} &  \multicolumn{3}{c}{Injuring} \\
         & P & R & F1 & P & R & F1 \\
        \midrule
        fine & $\mathbf{0.970}^*$ & $\mathbf{0.960}^*$ & $\mathbf{0.963}^*$ & $\mathbf{0.944}^*$ & $\mathbf{0.900}$ & $\mathbf{0.910}^*$ \\
        coarse & $0.826^{\transparent{0} *}$ & $0.930^{\transparent{0} *}$ & $0.861^{\transparent{0} *}$ & $0.799^{\transparent{0} *}$ & $0.889$ & $0.809^{\transparent{0} *}$ \\
        \bottomrule
    \end{tabular}}
    \label{tab:fine_grained}
\end{table}

 \paragraph{\textbf{Recommendation Accuracy}} As aforementioned, existing works on law articles recommendation works focus on recommending for the entire fact description paragraph. In order to compare the effects of recommendation methods with different granularities, we apply our MLMN to the coarse-grained recommendation task. We concatenated all facts in one paragraph together and predicted all relevant law articles as a whole. The input length of facts is expanded to 200. The performance comparison is shown in Table~\ref{tab:fine_grained}. The results are different from Table~\ref{tab:main_results} as the measures are all based on coarse-grained matching. We can see that the fine-grained recommendation improves the F1 score by more than 10\%, which verifies that refining granularity can make matching easier and more direct. 
 Note that in previous coarse-grained recommendation works~\cite{xu2020distinguish, YangJZL19}, their performances achieve the F1 score of 0.81 approximately which is slightly less than our best performance. The difference might be because that we use a single crime domain as the training unit which makes labels centralized and training becomes easier.

\begin{figure}
    \centering
    \begin{subfigure}[t]{0.8\linewidth}
        \centering
        \includegraphics[width=\linewidth]{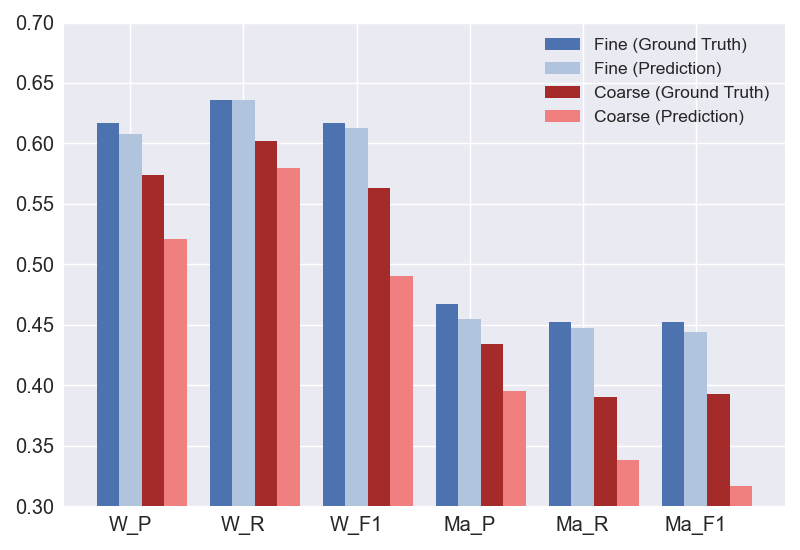}
        \caption{\small The crime of causing traffic accidents.}
    \end{subfigure}
    \begin{subfigure}[t]{0.8\linewidth}
        \centering
        \includegraphics[width=\linewidth]{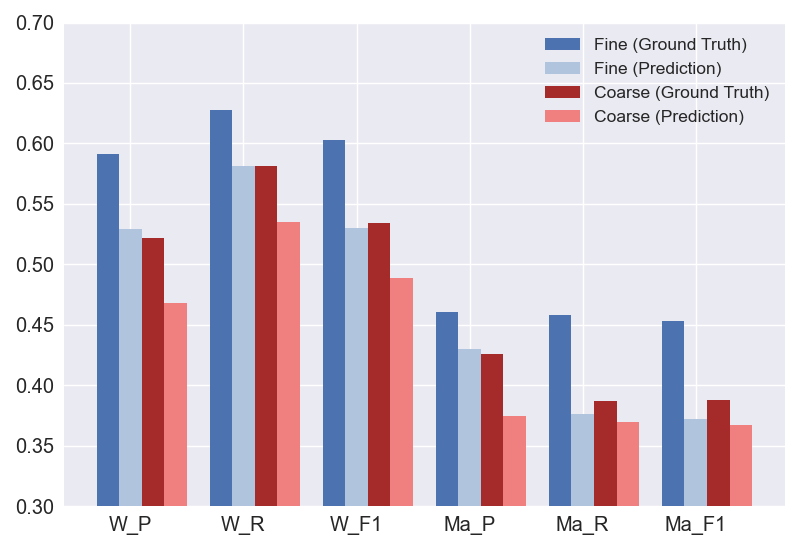}
        \caption{\small The crime of intentionally injuring.}
    \end{subfigure}
    \caption{\small Comparison of legal decision prediction with coarse / fine fact-article correspondences. ``W'' denotes ``Weighted'' and ``Ma'' denotes ``Macro'' in x-axis. ``Ground-truth'' means predicting the decision with ground-truth relevant law articles and ``prediction'' means with predicted ones.}
    \label{fig:downstream}
\end{figure}

\paragraph{\textbf{Decision Prediction Accuracy}}

In order to verify whether the fine-grained law article recommendation task can improve subsequent task of legal judgment prediction, we train a model to predict the term of penalty. We regard this task as a multi-class classification task which takes as input all facts and law articles involved in one case. In our designed model, we use a pre-trained Word2Vec model to generate the distributed representation, and then two Bi-LSTMs are used to encode individual facts and articles respectively. For the classifier which introduces fine-grained correspondence, we concatenate the representations of each fact with their related articles together, and then calculate their correspondence with the fully-connected neural network. At last, the correspondences that calculated based on each fact are added up to predict the final result. Because there may be more than one law articles related to one fact in reality, we sum all related articles up as the article's vector for subsequent calculations. 
On the contrary, for classifiers with coarse-grained fact-article correspondences, we only conduct one interaction between the entire fact paragraph and all law articles related to this case. The accumulated fact vector and law article vector are concatenated together for prediction directly. The predictions of these two classifiers are both conducted by the fully-connected neural network and Softmax. The experimental result is shown in Figure~\ref{fig:downstream}. We show results with both ground-truth related law articles and predicted ones. It can be seen that for different datasets, the classifier considering fine-grained correspondences is significantly better than on all metrics. We believe that clear correspondences can better simulate the process of judgment results prediction, thereby improving the performance. The gap between using ground-truth and predicted law articles is usually smaller with fine-grained annotations due to its high accuracy.

\subsection{Case Study}

\begin{figure}[h]
    \centering
    \includegraphics[width=\linewidth]{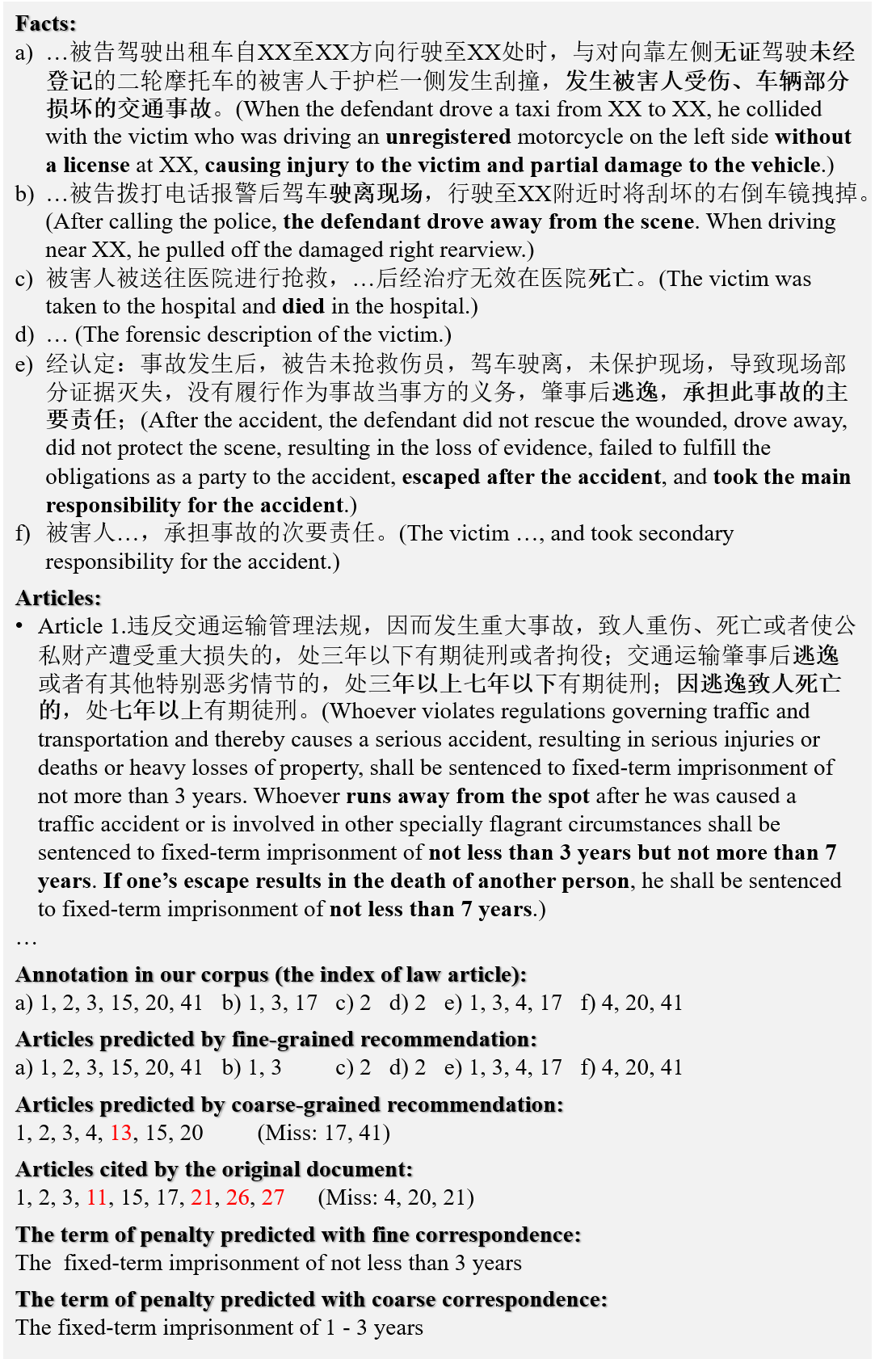}
    \caption{\small Example of law article recommendation and judgment decision prediction. The numbers marked in red are the mismatched articles compared with our annotation.}
    \label{fig:example}
\end{figure}

In this subsection, we provide an example of the application of fine-grained fact-article correspondence. As shown in Figure~\ref{fig:example}, there are six facts in this case and each fact can match different number of law articles. When conducting coarse-grained recommendation for all the six facts, there are one mistake and two missing citations for this case. Intuitively, the length of single fact is much shorter than the whole factual paragraph and the length of semantic dependencies to be captured is also shorter. It is much easier to capture the correspondence between a certain detail in the fact (escaped after the accident) and the article (Article 17 which defines the behavior of escaping), rather than matching with the whole fact sets. Besides, our corpus is imbalanced and constructed with cases in a single crime domain. If we use the entire factual paragraph as the training unit, some articles do not have any negative samples such as Article 1 which is a general law that can apply to almost all cases related with traffic accidents. For those articles with fewer positive samples (Article 41), the imbalanced issue is more severe for coarse-grained recommendation, which leads to the failure of prediction. 

There are also 4 mismatches and 3 missing citations for articles cited by the original document. These additional cited articles describe the behaviors of compensation (Article 11 \& 21) and insurance (Article 26 \& 27) which are relevant with other sections of this document and are not correspondent with the fact descriptions. The missing articles come from the judicial interpretation of criminal law (Article 4) and the Road Traffic Safety Law (Article 20 \&  21), which help provide clear definitions but have less impacts for final judgment decision, so they were omitted when writing the document. We remove the mismatched articles and include missing ones in our annotated corpus.

As for the prediction of judgment results, the correctness of prediction comes from the accurate recommendation of the key articles (Article 17 which defines the behavior as one kind of escaping). The judicial basis can be found in the premises of Article 1. In articles like Article 1, there are more than one premises in a single law article corresponding to multiple distinct facts, which increases the difficulty of prediction. When we introduce coarse-grained correspondences, the fact (b and e in the example) which matches one of the premises (escaping in this example) can be strengthened. When fact b and e are linked with Article 17, the model knows the case belongs to the category of escaping and makes the proper judgment result prediction (imprisonment of not less than 3 years). With only coarse-grained correspondence, it is difficult to identify the proper fact which defines the category of the behavior.

\section{Conclusion}

Automatic law article recommendation is an active research area. However, previous research can only make coarse-grained recommendation instead of pointing out the concrete fact-articles correspondences. In this paper, we construct a manually annotated corpus for fine-grained law article recommendation on two domains. We design a pipeline of careful annotation to remove missing, mismatched or incorrect articles in the extracted online legal documents. Based on this corpus, we propose a multi-level matching network which significantly outperforms other baselines in learning the correspondence. We also show that the fine-grained fact-article correspondences can not only improve the accuracy of recommendation, but also benefit other downstream tasks like predicting term of penalty. We hope our study can call for more attention on learning such fine-grained correspondence for a more interpretable and accurate AI system in law.

\bibliographystyle{ACM-Reference-Format}
\bibliography{sample-base}










\end{document}